\documentclass[sigconf]{aamas}

\usepackage{balance} 

\usepackage{subcaption}

\usepackage{graphicx} 

\setcopyright{ifaamas}
\acmConference[AAMAS '25]{Proc.\@ of the 24th International Conference
on Autonomous Agents and Multiagent Systems (AAMAS 2025)}{May 19 -- 23, 2025}
{Detroit, Michigan, USA}{A.~El~Fallah~Seghrouchni, Y.~Vorobeychik, S.~Das, A.~Nowe (eds.)}
\copyrightyear{2025}
\acmYear{2025}
\acmDOI{}
\acmPrice{}
\acmISBN{}

\acmSubmissionID{868}

\title[AAMAS-2025 Formatting Instructions]{Continual Reinforcement Learning via Autoencoder-Driven Task and New Environment Recognition
}

\author{Zeki Doruk Erden}
\affiliation{
  \institution{École Polytechnique Fédérale de Lausanne}
  \city{Lausanne}
  \country{Switzerland}}
\email{zeki.erden@epfl.ch}

\author{Donia Gasmi}
\affiliation{
  \institution{École Polytechnique Fédérale de Lausanne}
  \city{Lausanne}
  \country{Switzerland}}
\email{donia.gasmi@epfl.ch}

\author{Boi Faltings}
\affiliation{
  \institution{École Polytechnique Fédérale de Lausanne}
  \city{Lausanne}
  \country{Switzerland}}
\email{boi.faltings@epfl.ch}

\begin{abstract}

Continual learning for reinforcement learning agents remains a significant challenge, particularly in preserving and leveraging existing information without an external signal to indicate changes in tasks or environments. In this study, we explore the effectiveness of autoencoders in detecting new tasks and matching observed environments to previously encountered ones. Our approach integrates policy optimization with familiarity autoencoders within an end-to-end continual learning system. This system can recognize and learn new tasks or environments while preserving knowledge from earlier experiences and can selectively retrieve relevant knowledge when re-encountering a known environment. Initial results demonstrate successful continual learning without external signals to indicate task changes or reencounters, showing promise for this methodology.

\end{abstract}

\keywords{Continual learning, Reinforcement learning, Autoencoders}

\newcommand{\BibTeX}{\rm B\kern-.05em{\sc i\kern-.025em b}\kern-.08em\TeX}

\begin{document}


\pagestyle{fancy}
\fancyhead{}


\maketitle 


\section{Introduction}

Deep reinforcement learning \cite{li2017deep} has successfully tackled numerous challenges once deemed among the most difficult for intelligent agents, including tasks like intuition-based gaming \cite{silver2017mastering}, real-world robotics \cite{garaffa2021reinforcement}, and multi-agent systems \cite{zhang2021multi}. However, a major limitation of current systems is their inability to engage in continual learning—learning across dynamic environments without succumbing to destructive adaptation, where previously learned knowledge is lost. This stands in contrast to the abilities of human intelligence. For example, when children learn to ride a bicycle, they acquire the balancing skills necessary to maintain stability. These foundational skills are retained even as they go on to learn a variety of other tasks that also require balance, such as skateboarding or skiing, which do not destructively interfere with the skill of being balanced on a bike. Unfortunately, most approaches proposed to address the issue of destructive adaptation in continual learning in artificial agents rely on constraints within the problem domain, often requiring explicit storage and re-exposure to past data or assuming the presence of external task boundaries or signals for new tasks. These assumptions limit the systems to a constrained, partial form of continual learning.

In this work, we explore the use of autoencoders to enable continual learning in deep reinforcement learning agents, without relying on the limiting assumptions of past data replay or external task-boundary information. To achieve this, we employ an incrementally growing system design where the agent creates and utilizes a new neural network for each distinct environment, ensuring that learning in one environment does not interfere with others. Autoencoders are employed as a mechanism for detecting new environments or matching incoming observations to previously encountered ones. Our approach avoids explicit storage or replay of past samples and does not assume any designer-specified signals to inform the agent about environment changes, exposure to new environments, or which past environment it is currently facing.

\begin{figure}
     \centering
     \begin{subfigure}[t]{0.15\textwidth}
         \centering
         \includegraphics[width=\textwidth]{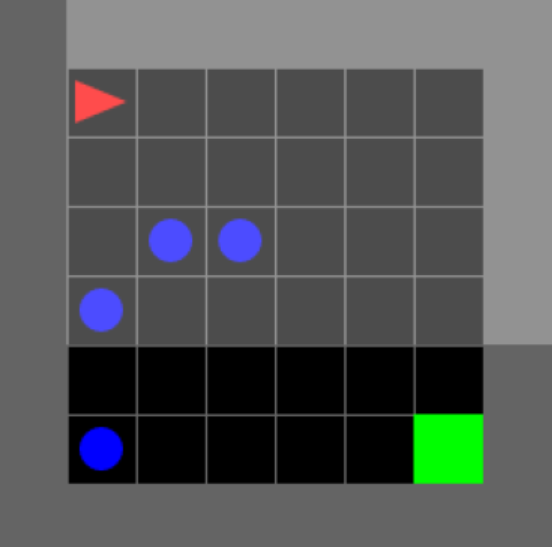}
         \caption{MG - DynamicObstacles}
         \label{fig:DynamicObstacles}
     \end{subfigure}
     \hfill
     \begin{subfigure}[t]{0.15\textwidth}
         \centering
         \includegraphics[width=\textwidth]{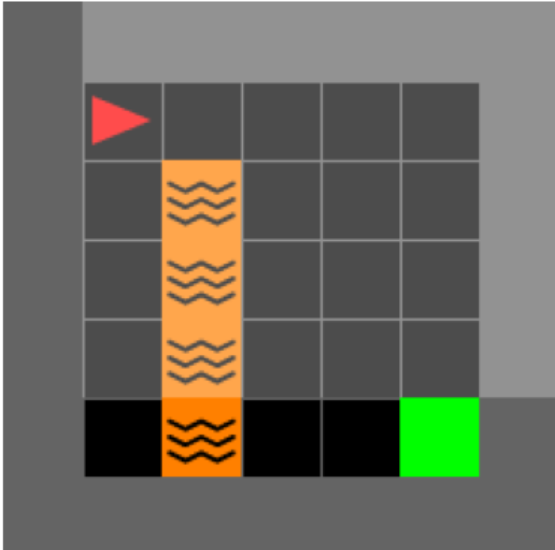}
         \caption{MG - LavaGap}
         \label{fig:LavaGap}
     \end{subfigure}
     \hfill
     \begin{subfigure}[t]{0.15\textwidth}
         \centering
         \includegraphics[width=\textwidth]{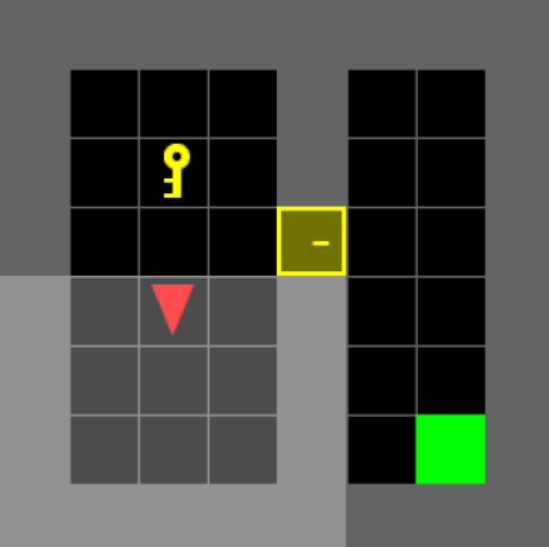}
         \caption{MG - DoorKey}
         \label{fig:DoorKey}
     \end{subfigure}
     \begin{subfigure}[t]{0.15\textwidth}
         \centering
         \includegraphics[width=\textwidth]{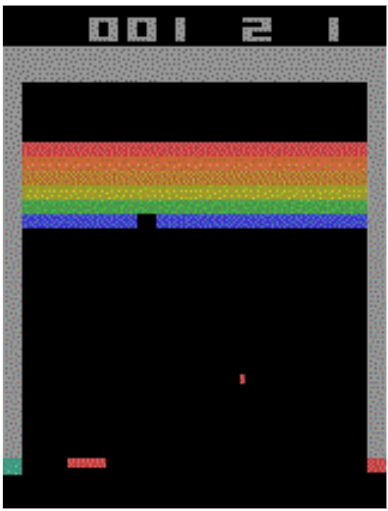}
         \caption{Atari - Breakout}
         \label{fig:Breakout}
     \end{subfigure}
     \hfill
     \begin{subfigure}[t]{0.15\textwidth}
         \centering
         \includegraphics[width=\textwidth]{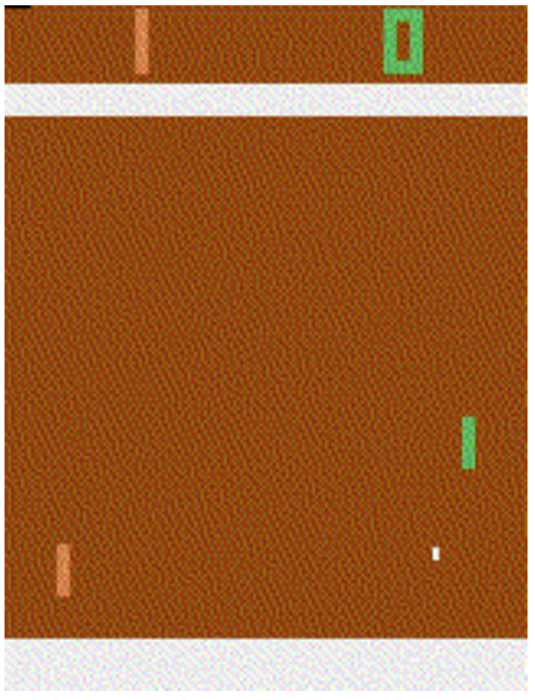}
         \caption{Atari - Pong}
         \label{fig:Pong}
     \end{subfigure}
     \hfill
     \begin{subfigure}[t]{0.15\textwidth}
         \centering
         \includegraphics[width=\textwidth]{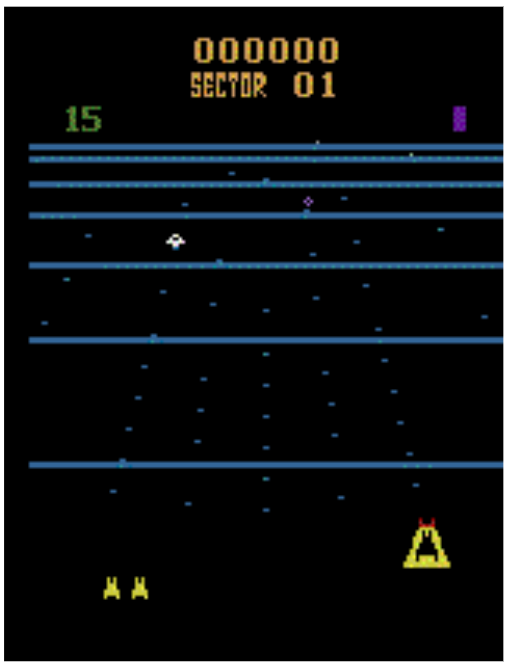}
         \caption{Atari - BeamRider}
         \label{fig:BeamRider}
     \end{subfigure}
    \caption{Our two experimental domains: Minigrid environment subtypes \cite{MinigridMiniworld23} (a-c) and Atari games \cite{towers2024gymnasium} (d-f) to be learned incrementally.}
    \label{fig:environments}
\end{figure}

We assess the efficacy of this approach in progressively learning multiple tasks across two experimental settings (Figure \ref{fig:environments}). The first setting is Minigrid \cite{MinigridMiniworld23}, a straightforward grid environment where an agent learns several tasks within a shared base environment. The second setting involves the Atari benchmarks \cite{towers2024gymnasium}, where the agent is tasked with successively mastering different games without forgetting its ability to play the earlier ones. These are well-known benchmarks for agent reinforcement learning \cite{mnih2013playing, chang2020decentralized}. In these domains, distinct environments and tasks necessitate a diverse set of skills, each with varying degrees of relevance depending on the task at hand, making sequential learning a natural approach. In both cases, we illustrate that a conventional reinforcement learning agent experiences a total erosion of prior knowledge, whereas our method enables continual learning without compromising performance.

\section{Related work}
\label{sec:related_work}

Continual learning, also known as incremental or lifelong learning, refers to the ability of AI systems to learn sequentially from an ongoing stream of tasks \cite{wang2024comprehensive, hadsell2020embracing}. A key challenge in continual learning is "catastrophic forgetting"—or, as we prefer to call it, "destructive adaptation"\footnote{We use the term "destructive adaptation" to avoid the anthropomorphic connotation of "catastrophic forgetting," which inaccurately suggests a gradual, human-like forgetting process. Instead, the phenomenon involves the active overwriting of past information, an issue that affects all adaptive systems.}. When new examples differ significantly from prior ones, they can overwrite previously learned knowledge in the network—a problem for which no reliable solution currently exists \cite{hadsell2020embracing, parisi2019continual, zollicoffernovelty}.

Fixed-capacity systems are inherently inadequate for addressing this problem: since neural networks encode information in a distributed fashion, the entire capacity is utilized by previous tasks. As a result, existing knowledge is eventually (and often immediately) overwritten when new tasks significantly diverge from earlier ones. On the other hand, methods that expand capacity face their own limitations. These systems cannot autonomously decide when to increase capacity, how to allocate new components to different tasks, or how to select the appropriate component when revisiting past tasks. For instance, in \cite{rusu2016progressive}, external signals are required both to recognize a new task and to specify which past task is being revisited. This limitation also affects methods that explicitly store information about previous task solutions, such as \cite{kirkpatrick2017overcoming}. Some extensions attempt to address these issues but still rely on task labels during adaptation and lack mechanisms for detecting new tasks, as seen in \cite{jacobson2022task}. Consequently, these methods fail to provide systems with fully autonomous continual adaptation capabilities.

Other approaches eliminate the need for explicit "task boundaries" by relying on the storage and replay of past data \cite{aljundi2019task, chaudhry2018efficient}, as noted in Table 1 of \cite{buzzega2020dark}. However, this strategy is impractical in many real-world settings, not only because it does away with one of the primary motivators of continual learning, but also due to the extensive memory demands over a system's lifetime, as well as concerns related to confidentiality and data protection.

The challenges of continual learning and destructive adaptation extend from the domain of supervised learning to reinforcement learning models as well \cite{abbas2023loss, khetarpal2022towards, abel2024definition}. Similarly, the solutions face comparable limitations, with reinforcement learning methods often relying on either the storage or replay of past data \cite{rolnick2019experience, traore2019discorl}, or the use of externally signaled task boundaries \cite{rusu2016progressive, schwarz2018progress}.

Our approach draws inspiration from the use of autoencoders for novelty detection \cite{pidhorskyi2018generative, del2022novelty, aljundi2017expert, jacobson2022task}, as well as the application of autoencoders in continual learning for supervised tasks by matching observed samples to previously learned "tasks" without needing external specification of the current task \cite{aljundi2017expert, jacobson2022task} (but still requiring external signification of an incoming new task). While novelty detection has been applied to reinforcement learning in some cases (e.g., \cite{zollicoffernovelty}), to our knowledge, it has not yet been used in conjunction with autoencoders to achieve full continual learning capabilities. This includes new task detection and prior task recognition, without the need for external signals or data replay, and without simplifying assumptions.

\section{Autoencoder-Driven Task and New Environment Recognition}
\label{sec:methods}

\subsection{Motivation and basis}

Our goal is to facilitate continual learning without destructive interference, avoiding explicit storage or replay of past samples, and without relying on external cues to indicate which task the agent is encountering or whether the current task is new or previously seen. In essence, we aim for the agent to preserve past knowledge intact, retrieve it as needed based on the demands of the current environment, and recognize when it should be acquiring new knowledge rather than reusing what it has already learned.

As outlined in Section \ref{sec:related_work}, any reliable method that ensures no loss of knowledge over an indefinite period and across an arbitrary number of tasks must incorporate mechanisms for expanding learning capacity. Accordingly, our design utilizes a system of multiple neural networks, each functioning as a policy network, added incrementally. Each policy network is tailored to a specific task or environment\footnote{We assume that a task change corresponds to an observable change in the environment. The generalization of our approach to handle changes in the environment's reward structure without observable changes is straightforward and covered in Section \ref{sec:conclusion}}. This approach aligns with established methods in the literature, where different tasks are associated with distinct, mostly or entirely independent neural networks \cite{rusu2016progressive, jacobson2022task}, facilitating the preservation of previously acquired knowledge.

\subsection{Task recognition}

Building on the foundation of multiple task-specific policy networks, our next goal is to implement a mechanism that can (1) assign the appropriate policy network based on the environment the agent encounters (retrieval) and (2) detect when the agent faces a previously unseen environment (new environment recognition), all without relying on external signals. To achieve this, we propose learning the features of the encountered environment during training, employing undercomplete autoencoders as a mainstream, efficient approach to do so \cite{michelucci2022introduction}. We leverage the ability of the autoencoder to reconstruct these features as an indicator of whether an incoming observation matches a previously learned environment, and as a quantification of the degree of match.\footnote{While simpler statistical methods, such as using averages and standard deviations per observation, can suffice in domains where tasks exhibit clear and broad differences, they fail in settings where most of the environment remains identical across tasks, with only subtle, sometimes conditional changes. The Minigrid \cite{MinigridMiniworld23} domain we experiment on is an example of this. The most general way to distinguish between environments, even in the presence of subtle variations, is to aim for a full reconstruction, capturing all conditional relationships.} Specifically, an undercomplete autoencoder (with a bottleneck layer smaller than the input/output dimensionality) learns to encode the key features necessary for reconstructing the original observation (e.g., an image) in a lower-dimensional space. This encoding is task-dependent, emphasizing features prominent to each training task. As a result, incoming data outside the distribution the autoencoder was trained on will manifest as imperfect reconstructions, signaling a new environment.

Our complete design is as follows (see Figure \ref{fig:Knowledge_Retrieval} for a summary, which is described in detail below):

\begin{figure*}[h!]
    \centering
    \includegraphics[width=0.8\textwidth]{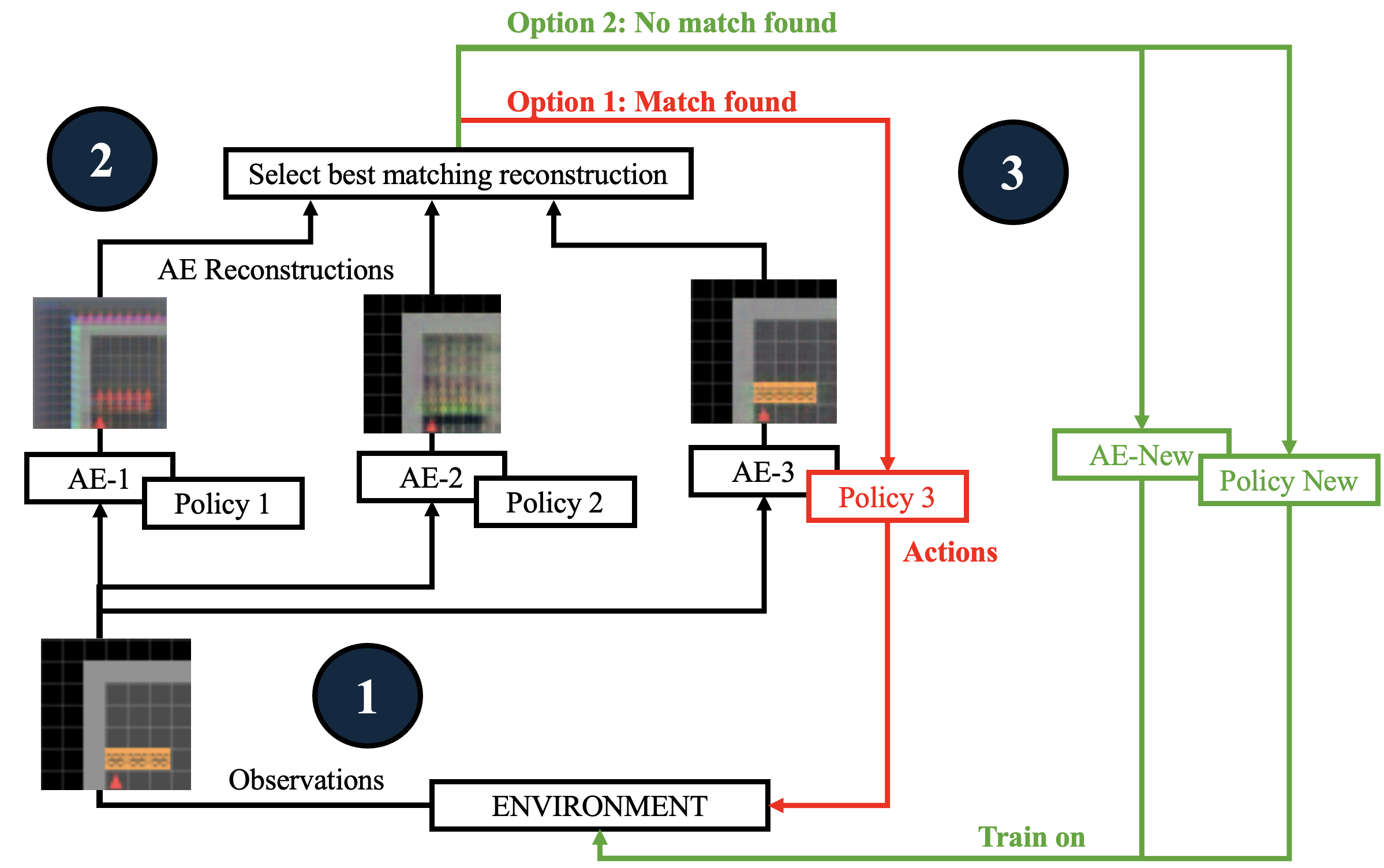}
    \caption{Overview of our system design. The system at any time is composed of a number of policy networks (three in this example) and an autoencoder (AE) associated with each of them. (1) The agent obtains observations from the environment (as images in our implementation). (2) The observations are passed as input to all autoencoders of all policy networks available for them to attempt reconstruction. The error on each autoencoder's reconstruction is computed. (3) If there are autoencoders whose reconstruction errors are below automatically estimated reconstruction thresholds (see main text), then the policy network associated with the autoencoder with the lowest reconstruction error is chosen (in this figure, that's Policy 3, meaning that AE-3 provided the lowest reconstruction error). If all reconstructions have errors above the threshold, this is interpreted as the observation of an unrecognized environment. A new policy network and a new associated autoencoder are created for training on this new task. (The illustrative reconstructions on this figure are actual in-operation outputs by our system, for tasks corresponding to Minigrid's Dynamic-Obstacles, Key-Door and Lava-Gap environments.)}
    \label{fig:Knowledge_Retrieval}
\end{figure*}

For each new environment, a policy network and a corresponding autoencoder are initialized. As the agent interacts with the environment and learns the policy, it simultaneously gathers observations. After the policy training is completed, these observations are used to train an environment-specific autoencoder. This autoencoder learns to capture the key features of the environment by minimizing the reconstruction error between the input observations and the reconstructed output.

\textit{Assumption:} We assume that a specific environment type remains available throughout training until the policy converges, or alternatively, that past training data is stored until the policy on a given environment reaches convergence (not to be confused with the storage of past data indefinitely for replay purposes, which we specifically avoid). This assumption does not impact the theoretical capabilities of our method; it merely defines what constitutes "one distinct environment/task." This assumption can be relaxed if we accept potential overfitting to different variations of the "same" environment (e.g., different room layouts in Minigrid Multiroom). In such cases, the agent would learn distinct policies for each unique variation (e.g., in Minigrid Multiroom, rooms with opposing doors versus neighboring doors would be treated as different environments). Alternatively, if the agent encounters frequently changing subtypes of an environment before convergence, all such variations would be grouped under the same policy network, and the associated autoencoder would adapt to encompass all subtypes (for instance, one policy could learn both Multiroom and Lava subtypes in Minigrid if these variants are repeatedly observed). These variations do not degrade performance or continual learning; they only affect the final configuration of the policy and autoencoder networks.

Once the autoencoder is trained, an estimation of a \textit{reconstruction error threshold} is computed based on the distribution of reconstruction errors during training, \textit{per task}. We do this by using a batch-wise averaging approach. First, the reconstruction errors are calculated for batches of validation observations. A Gaussian distribution is then fitted to the batch-wise average reconstruction errors. The threshold is determined by calculating the value that corresponds to a specified confidence level (e.g. 99\%) using the cumulative distribution function (CDF) of the fitted distribution. This threshold ensures that reconstruction errors exceeding it are flagged as indicative of a novel environment. This approach enables us to dynamically determine the expected reconstruction performance for different environment types—an important consideration, as our preliminary experiments revealed that typical reconstruction errors vary significantly across tasks. As a result, a fixed reconstruction threshold cannot be applied uniformly.


To recognize novelty, the collected observations are passed through a set of trained autoencoders from previously encountered environments. If all the autoencoders produce reconstruction errors above their respective estimated thresholds (see above), the environment is classified as novel, prompting the training of a new autoencoder and policy network for the new environment. However, if one or more autoencoders reconstruct the observations with an error below their thresholds, the environment is recognized as familiar. In such cases, the autoencoder with the smallest reconstruction error is identified as the best match, and its corresponding policy network is used to continue interacting with the environment.

\subsection{Summary and illustrative example}

Figure \ref{fig:Knowledge_Retrieval} illustrates the agent’s novelty recognition process using autoencoders, exemplified with some reconstructions from our experiments in Minigrid environment variants. Here, the agent’s knowledge includes three environments: LavaGap, DoorKey, and DynamicObstacles. The agent’s current observation (from the LavaGap environment) is passed through all three autoencoders for the three environment types. In the first alternative outcome (Option 1 in Figure \ref{fig:Knowledge_Retrieval}), Autoencoder 3 (trained on LavaGap) yields the lowest reconstruction error, below its corresponding threshold, allowing the agent to identify the environment as the one that its policy, Policy 3, was trained on (again, LavaGap); hence it retrieves this policy model for interaction with the environment. If, on the other hand, none of the autoencoders match (i.e. all reconstruction errors are above their thresholds - Option 2 in Figure  \ref{fig:Knowledge_Retrieval}), the agent recognizes the environment as novel. For example, if a fourth environment is introduced, the agent would start training a new PPO model and a new autoencoder, and estimate a threshold for the new task, adding it to its knowledge for future encounters. In that manner, the agent can learn an arbitrary number of future tasks, without destroying knowledge about any of the past tasks.

Finally, we would like to note that while we used this framework of associated autoencoders in conjunction with standard neural networks as basis, the method has no conflict with and can just as well be used with other continual learning frameworks as basis instead - particularly some established methods that require task boundaries provide good candidates for such an integration \cite{rusu2016progressive, kirkpatrick2017overcoming}.

\section{Experimental setup}

\subsection{Implementational details of the system}

\textit{Policy:} For the implementation of Proximal Policy Optimization (PPO), we used Stable Baselines3. The default settings were kept. We used CnnPolicy due to the image-based nature of the environments. The optimizer used is Adam, with a learning rate of $3e-4$.

\textit{Autoencoder:} We trained a convolutional autoencoder to reconstruct environment observations. The convolutional autoencoder for MiniGrid has input images of shape (7, 7, 3), uses two Conv2D layers in the encoder (16 and 8 filters, 3x3 kernels, ReLU activation), each followed by 2x2 MaxPooling. The decoder employs two Conv2DTranspose layers for upsampling, followed by a final Conv2D layer (3 filters, sigmoid activation) to reconstruct the original input. For Atari input images have shape (84, 84, 1) (downscaled and grayscaled from 210x160x3), uses two Conv2D layers in the encoder (16 and 8 filters, 3x3 kernels, ReLU activation), each followed by 2x2 MaxPooling to downsample. The decoder uses two Conv2DTranspose layers for upsampling, followed by a final Conv2D layer (1 filter, sigmoid activation) to reconstruct the grayscale input. The models were optimized using the Adam optimizer and a binary cross-entropy loss function, with early stopping based on validation loss to prevent overfitting. The autoencoders were trained for up to 100 epochs with a batch size of 64, using 20\% of the data for validation. Training and validation loss were monitored and plotted to assess model performance. For the desired confidence level to determine autoencoder reconstruction thresholds, we used 90\% in Minigrid experiments and 99\% in Atari experiments (we used a stricter threshold for Atari since they are visually more similar).

We classify the observations (or identify them as new) and select a policy network only once per episode, at the outset. This approach assumes that there is no change in the environment during a single episode, but only between episodes. However, this assumption can be relaxed without altering our system design if the environment is non-episodic.
\subsection{Experiment details}

In our experiments, we use Minigrid \cite{MinigridMiniworld23} and Atari environments provided by the OpenAI/Gym toolkit \cite{towers2024gymnasium}. We test with three environment subtypes (i.e. tasks) from each domain: In Minigrid, we use the variants Dynamic-Obstacles-8x8-v0 (Task T1), LavaGapS7-v0 (Task T2), and DoorKey-8x8-v0 (Task T3). In Atari, we test with games Breakout (Task T1), Pong (Task T2), and BeamRider (Task T3), all v5. All our results with Minigrid are averages of 8 runs, and with Atari they are averages of 3 runs (the lower number of runs was due to the computational cost of learning Atari environments due to their high dimensionality). In all our experiments, the agent gets observations as images (RGB for Minigrid, grayscale for Atari). Since reward magnitudes differ greatly across tasks, reported performances are normalized to the range of $[0,1]$, with normalization limits determined by the highest and lowest rewards obtained by the agent in the corresponding environment.

We conduct our experiments with an agent implementing our design, referred to as the "AE-CL Agent" (short for "autoencoder continual learning"), alongside a vanilla agent that utilizes a single policy network for comparison. We do not include any additional baseline comparisons, as we are unaware of any comparable continual reinforcement learning methods that can achieve continual learning without the external specification of task IDs or new task information, or without replaying past samples. Comparing our approach to methods that rely on these constraints would not provide meaningful insights, as our primary objective is to showcase the capability of our design to function without such constraints. Moreover, we do not present this method as an alternative to existing techniques that operate under these assumptions (e.g. \cite{rusu2016progressive, kirkpatrick2017overcoming}); rather, it has the potential to work in conjunction with them—especially those requiring task boundaries without automatic detection. Nothing in our design prevents an integration with such existing approaches in place of standard neural networks as policy basis as we use them, as discussed in detail on Sections \ref{sec:methods} and \ref{sec:conclusion}.

\subsubsection{Learning flow 1: Retrospective Performance}

In this learning flow, we specifically demonstrate the agent's knowledge preservation and retrieval performance on previous tasks. The agent is sequentially exposed to each task, and after learning one task, it is tested for average performance across all the tasks that it was exposed to until that point. Specifically, the training and testing process unfolds as follows:

\begin{enumerate}
    \item The agent is trained on the first task (T1). After learning T1, the agent was tested on 30 episodes of T1 only.
    \item The agent is trained on the second task (T2). After learning T2, the agent was evaluated on 60 episodes, split equally between T1 and T2. The transitions between episodes were random, with the environment selection (T1 or T2) chosen randomly for each episode.
    \item The agent is trained on the second task (T3), and following that the agent was tested across 90 episodes where the environments (T1, T2, T3) were randomly selected. One-third of the episodes were from each task (T1, T2, and T3).
\end{enumerate}

\begin{figure}[h!]
    \centering
    \includegraphics[width=0.5\textwidth]{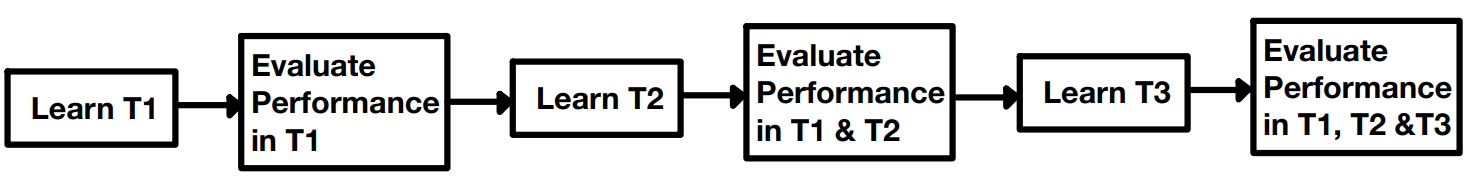}
    \caption{Learning Flow 1 (Retrospective performance)}
    \label{fig:Experiment1 Flow Diagram}
\end{figure}

This flow is visualized on Figure \ref{fig:Experiment1 Flow Diagram}. We would like to reiterate that at no point during the training process do we inform the agent of a task change, the introduction of a new task, or which of the previous tasks it is currently encountering.

\subsubsection{Learning flow 2: Ongoing Performance}

\begin{figure}[h!]
    \centering
    \includegraphics[width=0.5\textwidth]{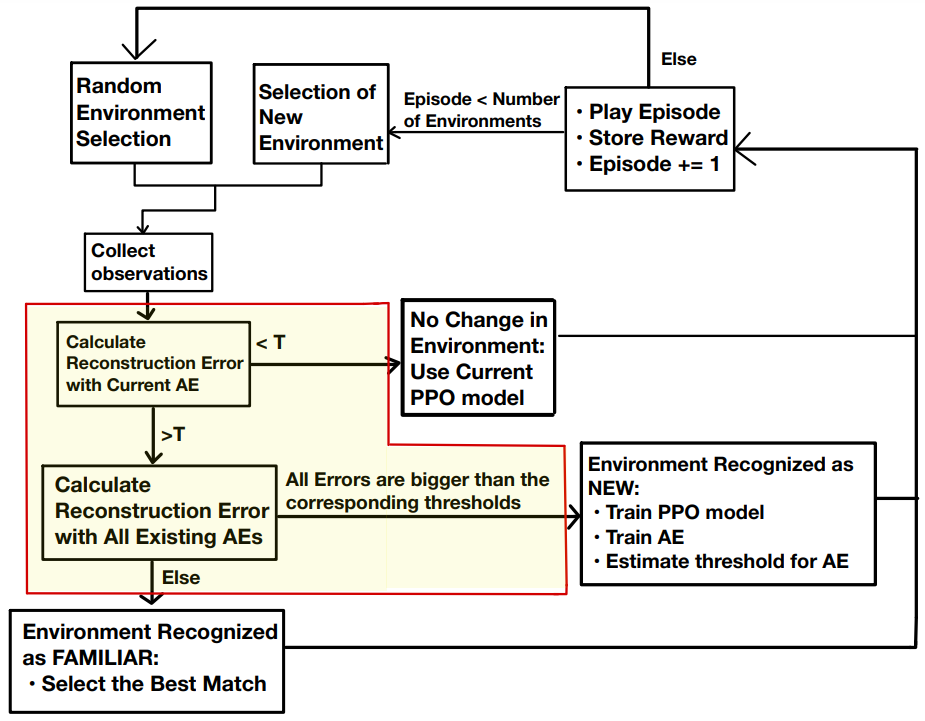}
    \caption{Learning Flow 2 (Ongoing performance)}
    \label{fig:Experiment2 Flow Diagram}
\end{figure}

In this learning flow (visualized in Figure \ref{fig:Experiment2 Flow Diagram}), we test the AE-CL agent's performance in a more natural, ongoing operation. We provide the agent with a distinct environment at each step, and we track the net performance of the agent across all these steps, without distinction across tasks; arguably simulating a more natural scenario compared to the retrospective evaluations in the first training flow. As congruent with our assumption (Section \ref{sec:methods}), we assume that an environment remains accessible until the convergence of the agent on that environment. To test continual learning performance, both AE-CL and Vanilla agents are trained on a given task only at their first exposure to this task (note that AE-CL does this automatically, while for Vanilla agent this was done manually).



\section{Results and discussion}

\subsection{Number of distinct poliy-autoencoder pairs learned}

In all our experiments we saw that the agent learned precisely three policy-autoencoder pairs, without any tasks missed or without any multiple unnecessary policy-autoencoder pairs for a single task.

\subsection{Retrospective performance}

Tables \ref{tab:minigrid_exp1} and \ref{tab:atari_exp1} present the retrospective performance evaluations for the Minigrid and Atari environments, respectively. In all instances, the AE-CL agent (our proposed design) demonstrates consistent average performance across all tasks introduced up to that point. For example, in the Atari environment (Table \ref{tab:atari_exp1}), the agent achieves an average normalized reward of 0.945 across Tasks 1, 2, and 3 combined (after being trained on Task 3), which is nearly identical to its original single-task performance of 0.951 on Task 1, despite not being retrained on Tasks 1 and 2. In contrast, the Vanilla agent experiences destructive adaptation, losing all knowledge of previous tasks upon the introduction of a new one, resulting in a normalized reward of approximately 1/X after task index X (i.e. achieving a reward close to 1.0 on the latest task but around 0 for all prior tasks).

\begin{table}[]
    \centering
    \begin{tabular}{c|c|c|c}
         \textbf{Agent} & \textbf{Task 1} & \textbf{Task 2} & \textbf{Task 3} \\
         \hline
        AE-CL & 0.901 (0.013) & 0.887 (0.007) & 0.907 (0.004) \\
        Vanilla & 0.907 (0.013) & 0.430 (0.003) & 0.324 (0.001)
    \end{tabular}
    \caption{Retrospective performances of AE-CL and Vanilla RL agents on Minigrid. Displayed values under Task X are average rewards (normalized) across all tasks up to and including Task X (e.g. the performance under Task 2 is the retrospective performance on Task 1 and 2 combined). Tasks 1, 2 and 3 are DynamicObstacles, LavaGap and DoorKey environments respectively. Results are averaged over 8 independent runs, inside parentheses are standard deviations.}
    \label{tab:minigrid_exp1}
\end{table}

\begin{table}[]
    \centering
    \begin{tabular}{c|c|c|c}
         \textbf{Agent} & \textbf{Task 1} & \textbf{Task 2} & \textbf{Task 3} \\
         \hline
        AE-CL & 0.951 (0.026) & 0.939 (0.010) & 0.945 (0.007) \\
        Vanilla & 0.947 (0.026) & 0.469 (0.006) & 0.310 (0.002)
    \end{tabular}
    \caption{Retrospective performances of AE-CL and Vanilla RL agents on Atari. Displayed values under Task X are average rewards (normalized) across all tasks up to and including Task X. Tasks 1, 2 and 3 are Breakout, Pong and Beamrider respectively. Results are averaged over 3 independent runs, inside parentheses are standard deviations.}
    \label{tab:atari_exp1}
\end{table}

\begin{figure}[h!]
    \centering
    \includegraphics[width=0.4\textwidth]{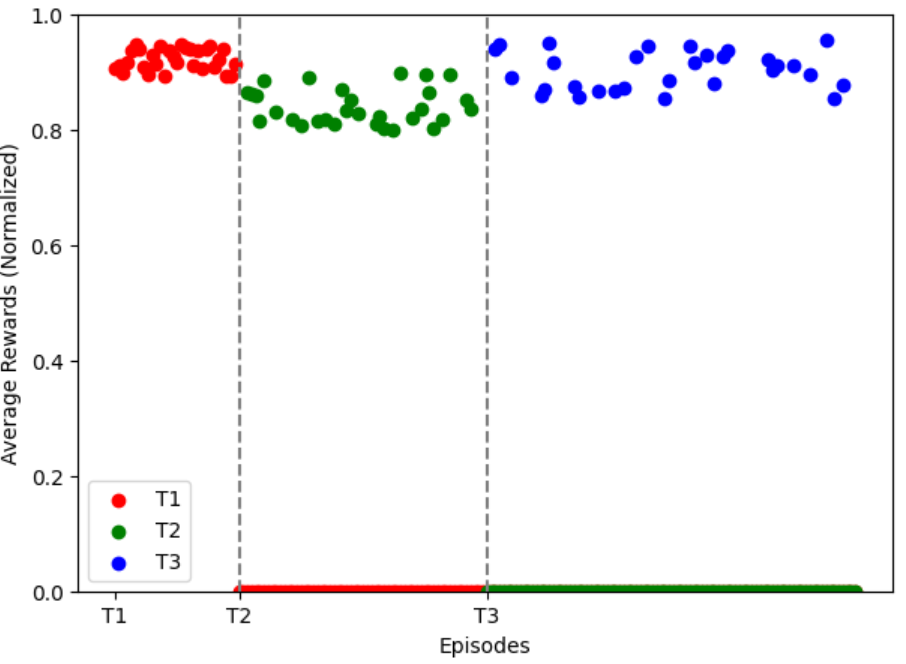}
    \caption{Normalized rewards obtained by Vanilla agent during retrospective performance evaluation at the three tasks. Labels T1, T2 and T3 on x-axis signify the time of training on the corresponding task, followed by subsequent evaluation on tasks up to that point. The plot shows clear destructive adaptation as performance in prior tasks are not retained.}
    \label{fig:Retrospective_Vanilla_Agent}
\end{figure}

\begin{figure}[h!]
    \centering
    \includegraphics[width=0.4\textwidth]{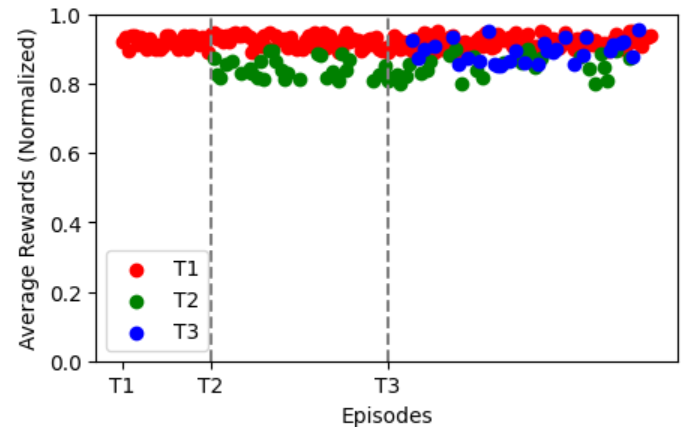}
    \caption{Normalized rewards obtained by AE-CL agent during retrospective performance evaluation at the three tasks. Labels T1, T2 and T3 on x-axis signify the time of training on the corresponding task, followed by subsequent evaluation on tasks up to that point. Performance retained without any loss upon the introduction of new tasks.}
    \label{fig:Retrospective_Continual_Agent}
\end{figure}

Figure \ref{fig:Retrospective_Vanilla_Agent} gives a different view on destructive adaptation of the Vanilla Agent, where the performance on any prior task is seen to have decayed to 0 after being trained on a new task. In contrast, the AE-CL agent (Figure \ref{fig:Retrospective_Continual_Agent}) retains original performance in prior environments after being trained on new ones, showing no sign of destructive adaptation.

The results on retrospective performances learning flow shows that the AE-CL agent can accurately recognize new tasks and correctly assign observations from previously-encountered tasks to their corresponding policy-autoencoder pairs, hence realizing continual learning without any external task change or new task signals.

\subsection{Ongoing performance}

\begin{figure}[h!]
    \centering
    \includegraphics[width=0.4\textwidth]{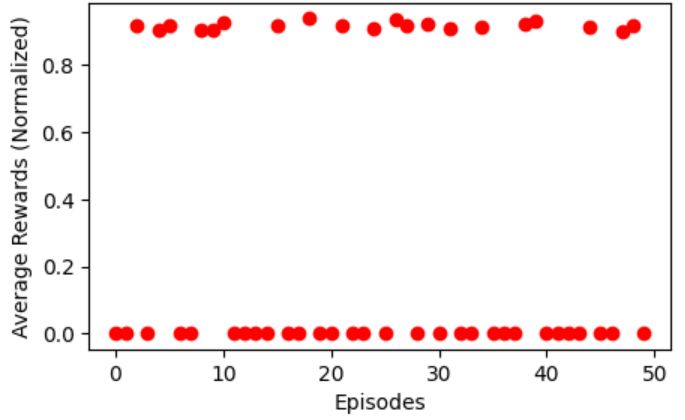}
    \caption{Normalized rewards obtained by the Vanilla agent on Minigrid across 50 episodes during ongoing performance evaluation. The environment subtype is chosen randomly at each episode.}
    \label{fig:Ongoing_performance_vanilla}
\end{figure}

\begin{figure}[h!]
    \centering
    \includegraphics[width=0.4\textwidth]{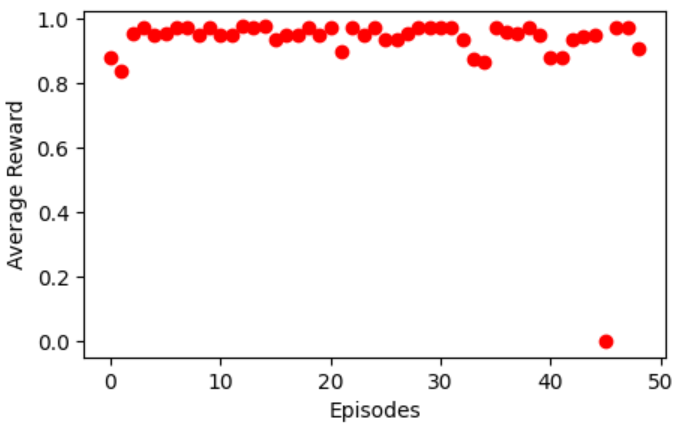}
    \caption{Normalized rewards obtained by the AE-CL agent on Minigrid across 50 episodes during ongoing performance evaluation. The environment subtype is chosen randomly at each episode.}
    \label{fig:Ongoing_performance}
\end{figure}

Table \ref{tab:Ongoing_performance} presents the average performance of the AE-CL agent during ongoing performance evaluations across three tasks, demonstrating high average performance across trials after learning three policy-autoencoder pairs during its initial encounters with each environment (as discussed at the beginning of this section). In contrast, Vanilla agent once again shows a performance close to 1/3rd of maximum performance, demonstrating that the knowledge of all tasks but the latest-trained one are lost. Figures \ref{fig:Ongoing_performance_vanilla} and \ref{fig:Ongoing_performance} illustrate this for 50 successive episodes randomly selected from our Minigrid experiments. The Vanilla agent (Figure \ref{fig:Ongoing_performance_vanilla}) performs well in approximately 1/3rd of episodes while completely failing at the rest. AE-CL agent (Figure \ref{fig:Ongoing_performance}), on the other hand,  achieves near-optimal performance across all episodes except for one instance (episode 45), where the environment subtype was correctly identified as LavaGap, but the agent failed due to an incorrect action taken by the policy network (which is not unlikely in LavaGap environment, since falling into lava cells by mistake results in immediate episode termination). This indicates that the AE-CL agent is capable of accurately detecting and classifying the environments it encounters in an ongoing manner, retrieving relevant past knowledge, and performing effectively based on that retrieval.

\begin{table}[]
    \centering
    \begin{tabular}{c|c|c}
        \textbf{Agent} & \textbf{AE-CL}  & \textbf{Vanilla} \\
        \hline
        Minigrid & 0.895 (0.005) & 0.283 (0.080) \\
        Atari & 0.942 (0.008) & 0.319 (0.080) \\
    \end{tabular}
    \caption{Net average rewards (normalized) of AE-CL and Vanilla agents during ongoing evaluation. Results are averaged over 8 independent runs for Minigrid and 3 runs for Atari, inside parentheses are standard deviations.}
    \label{tab:Ongoing_performance}
\end{table}

\section{Conclusions}
\label{sec:conclusion}

In this study, we proposed a training flow for continual learning based on dynamically growing system capacity to avoid destructive adaptation, combined with autoencoders for task assignment and new environment detection. We showed that this simple design can successfully detect new environments and accurately assign observations to previously-encountered environments if they provide a match, hence enabling continual learning without requiring external signals for neither of these tasks, and without requiring storage of past training data. Our system design can be used either as it is presented in this paper (with vanilla neural networks) or it can be used in conjunction with other methods that typically assume the existence of task boundaries, such as Progressive Neural Networks \cite{rusu2016progressive} or Elastic Weight Consolidation \cite{kirkpatrick2017overcoming}.

\subsection{Limitations}

Our current design is tailored for detecting changes in the environment, signaled by shifts in the observation distribution, which affects the autoencoder’s ability to accurately reconstruct the original observation. The key constraint here, with respect to a fully generalized continual learning setup, is the assumption that each new learning task is linked to a novel environment. While this assumption holds for many problem scenarios, in principle, a "task change" could also be defined by a shift in the reward structure within the same environment. Our framework can easily be adapted to accommodate such cases by extending the autoencoder’s inputs and outputs to include the rewards obtained by the agent. This would allow the system to detect changes in both the reward structure and the underlying observations, thereby generalizing its continual learning capabilities.

While our method enables the learning of an arbitrary number of tasks without destructive adaptation due to its ability to grow capacity as needed, this also implies that memory requirements may become substantial for very long-lived agents. Each environment/task is represented by a distinct policy-autoencoder network, which presents a shared limitation across continual learning methods that increase capacity \cite{rusu2016progressive, jacobson2022task}. Approaches that do not expand capacity (e.g., \cite{kirkpatrick2017overcoming}) circumvent this limitation; however, as discussed, they cannot continue learning indefinitely long sequences of tasks due to their finite capacity. We believe that an effective solution to this limitation lies in mechanisms that can incrementally add partial capacity (such as network components, neurons, or layers, rather than entire neural networks), which would help control the increase in complexity, ideally following a logarithmic trend as the capacity required decreases with the agent's expanding experience. This concept also ties into our discussion of transfer learning in the subsequent subsection.

\subsection{Future work}

A key motivation for continual learning is knowledge transfer, where knowledge gained from past tasks is used to enhance performance on subsequent ones \cite{zhuang2020comprehensive}. Our current system design and experiments do not yet incorporate this possibility, as the agent begins learning a new policy network and a new autoencoder with each new task it encounters. However, the design can be extended to integrate existing transfer learning techniques, allowing it to leverage prior knowledge from previous environments. One straightforward approach would be to initialize the policy network for a new task using the network associated with the autoencoder that best matched the new observation during reconstruction. We conducted preliminary experiments with this transfer method but did not report the results, as we observed no significant impact on performance or training progression. This may be due to a lack of similarity across environments in our experimental domain, where it is more efficient to learn a new network from scratch than to reuse an existing one. In scenarios with greater overlap between tasks or environments, this transfer strategy could have a more pronounced effect. Alternatively, it is possible that such a simple transfer scheme is inherently ineffective, and more advanced transfer learning techniques may be needed to effectively harness prior knowledge and further enhance training - our method, currently built on standard neural networks, can be easily extended to incorporate such alternative transfer learning approaches.

While our primary emphasis throughout this paper has been on the quantitative performance of continual learning, it is important to recognize that our method not only facilitates continual learning without destructive adaptation but also decouples and represents different tasks (currently represented as distinct environment subtypes, but not necessarily limited to that, as previously discussed) as isolated behavioral subunits. This aspect may be of particular interest to researchers engaged in Hierarchical Reinforcement Learning \cite{pateria2021hierarchical}, which, among other objectives, seeks to represent distinct components of an overall behavioral pattern as separate entities. Although we did not explore this dimension of our system in this study, we believe it presents a promising avenue for future research for those interested in this field.





\end{document}